\documentclass[final]{cvpr}

\usepackage{times}
\usepackage{epsfig}
\usepackage{graphicx}
\usepackage{amsmath}
\usepackage{amssymb}

\usepackage{amsfonts}
\usepackage{bbm}
\usepackage[ruled,vlined,noend]{algorithm2e}

\usepackage[pagebackref=true,breaklinks=true,colorlinks,bookmarks=false]{hyperref}

\usepackage[capitalise]{cleveref}

\newtheorem{definition}{Definition}
\crefname{definition}{definition}{definitions}
\Crefname{definition}{Definition}{Definitions}

\newtheorem{lemma}[definition]{Lemma}
\crefname{lemma}{lemma}{lemmas}
\Crefname{lemma}{Lemma}{Lemmas}

\newtheorem{proposition}[definition]{Proposition}
\crefname{proposition}{proposition}{propositions}
\Crefname{proposition}{Proposition}{Propositions}

\newtheorem{remark}[definition]{Remark}
\crefname{remark}{remark}{remarks}
\Crefname{remark}{Remark}{Remarks}

\DeclareMathOperator{\relu}{ReLU}

\def\cvprPaperID{} 
\def\confYear{CVPR 2021}

\begin{document}

\title{Neural Response Interpretation through the Lens of Critical Pathways}

\author{Ashkan Khakzar\textsuperscript{1}, 
Soroosh Baselizadeh\textsuperscript{1}, 
Saurabh Khanduja\textsuperscript{1} \\ 
Christian Rupprecht\textsuperscript{2}, 
Seong Tae Kim\textsuperscript{1}, 
Nassir Navab\textsuperscript{1}\\
{\tt\small ashkan.khakzar@tum.de}\\
\\
\textsuperscript{1}CAMP @ Technical University of Munich\\
\textsuperscript{2}VGG @ University of Oxford\\
}


\maketitle

\begin{abstract}
Is critical input information encoded in specific sparse pathways within the neural network?
In this work, we discuss the problem of identifying these critical pathways and subsequently leverage them for interpreting the network's response to an input.
The pruning objective
--- selecting the smallest group of neurons for which the response remains equivalent to the original network --- 
has been previously proposed for identifying critical pathways.
We demonstrate that sparse pathways derived from pruning do not necessarily encode critical input information.
To ensure sparse pathways include critical fragments of the encoded input information, we propose pathway selection via neurons' contribution to the response.
We proceed to explain how critical pathways can reveal critical input features. We prove that pathways selected via neuron contribution are locally linear (in an $\ell_{2}$-ball), a property that we use for proposing a feature attribution method: ``pathway gradient''.
We validate our interpretation method using mainstream evaluation experiments.
The validation of pathway gradient interpretation method further confirms that selected pathways using neuron contributions correspond to critical input features. The code\footnote{\url{https://github.com/CAMP-eXplain-AI/PathwayGrad}} \footnote{\url{https://github.com/CAMP-eXplain-AI/RoarTorch}}
 is publicly available.

\end{abstract}

\section{Introduction}

Understanding the rationale behind the response of a neural network is of considerable significance. Such transparency is required for adoption and safe deployment in mission-critical domains.
Interpreting the response also helps in debugging and designing neural networks, and quenches the intellectual curiosity over how neural networks function \cite{goodman2017european,Ribeiro2016,kim2018interpretability, khakzar2019learning,wang2018interpret,fong2017interpretable,selvaraju2017grad}. 

What insights can we acquire about the underpinnings of a neural network's response by putting the networks under the microscope and analyzing neurons and pathways? By "\emph{pathway}", we refer to a union of paths (equivalently a sub-network) that connect the input to the output. Discovering to what patterns neurons correspond --- also known as neural decoding in computational neuroscience \cite{dayan2001theoretical} --- has revealed human interpretable concepts encoded in neurons of artificial neural networks, \eg curve and circle detectors \cite{olah2017feature,zeiler2014visualizing}. Analyzing the neural pathways has recently revealed human interpretable \emph{connections} between concepts encoded within each neuron on the pathway, \eg circles being assembled from curves \cite{olah2020zoom}. In this work we discuss pathways responsible for the network's response given a \emph{specific input}, but how can we identify these pathways?

Deep rectified neural networks encode the input information using a sparse set of active neurons \cite{Glorot2011}, and their inference can be deemed as a pursuit algorithm for sparse coding ~\cite{papyan2017convolutional,sulam2018multilayer}. Such sparse coding of information is akin to how biological neurons encode information in the brain \cite{olshausen2004sparse,hahnloser1998piecewise}. Yu \etal \cite{yu2018distilling} reported that the pathways of active neurons in artificial neural networks overlap significantly for inputs of a given class. Recently, \cite{wang2018interpret} proposed using the pruning objective and knowledge distillation \cite{Hinton06} to show that significantly higher levels of sparsity ($\sim$87\% for VGG-16 \cite{simonyan2014very} on ImageNet \cite{deng2009imagenet}) can be achieved while keeping the prediction intact. These highly sparse pathways are reported as the critical paths and are shown to be different for inputs of different classes and adversarial inputs \cite{Qiu2019profilepath, wang2018interpret,yu2018distilling}. 

We first investigate, whether these highly sparse pathways derived from the pruning objective indeed encode critical input features. \emph{We show that the pruning objective has solutions that are not critical pathways}, even though they have the same response as the original network. To illustrate \emph{how} the pruning objective can result in such pathways, we construct a pathological greedy pruning algorithm that by design searches for irrelevant pathways while satisfying the pruning objective. Furthermore, we analyze the pathways selected by distillation guided routing \cite{wang2018interpret} and observe a similar phenomenon. We also use feature visualization ~\cite{olah2017feature,mahendran2015understanding} to \emph{decode} and semantically analyze the pathways.
If these pathways do not encode critical input features, how can we find such pathways?
Numerous works have studied the importance of individual neurons for the neural response, and how each neuron encodes information specific to one or a subset of classes \cite{Zhou2018,bau2017network,Morcos2018,olah2017feature}. It is therefore intuitive that selected sparse pathways should encompass \emph{important/critical} neurons for the corresponding response. We thus investigate selecting pathways based on neuron contributions as opposed to the pruning objective. In order to compute the importance of neurons, we use notions of marginal contribution and the \emph{Shapley} value \cite{shapley1953value,Lundberg2017,Ancona2019,sundararajan2020many,Zintgraf2019}.
The first section of the work is devoted to the discussion of critical pathways.
 
We proceed to answer how critical pathways can help us interpret the response of the network. We prove that in rectified neural networks, pathways selected by neuron contributions are locally linear. 
We leverage this property and propose an input feature attribution methodology which we refer to as "pathway gradient".
We evaluate our attribution methodology with input degradation \cite{Samek2017}, sanity checks \cite{Adebayo2018}, and Remove-and-Retrain (ROAR) \cite{hooker2019benchmark} on Cifar10 \cite{krizhevsky2009learning}, Bridsnap \cite{berg-birdsnap-cvpr2014}, and ImageNet \cite{deng2009imagenet} datasets. By validating our attribution methodology, we also validate that selected pathways using neuron contributions indeed correspond to critical input features.
In summary, the main contributions of the paper are:
\\
\\
$\bullet$ We show that the pruning objective does not necessarily extract critical pathways. We illustrate how the pruning can fail by proposing a pathological greedy algorithm that by design searches for irrelevant pathways. Subsequently, we propose selecting pathways based on neuron contributions instead.
\\
$\bullet$ We prove that critical pathways selected by neuron contributions are locally linear ($\ell_{2}$-ball) in rectified networks. Using local linearity, we propose a feature attribution approach, "pathway gradient", that reveals input features associated with features encoded in the critical pathways.
\\
$\bullet$ We empirically show that computing contribution (approximated Shapley value) of \textit{neurons} rather than input pixels, improves input feature attribution.

\section{Background and related work}\label{sec:related_work}

\textbf{Feature visualization / Neural decoding:}
This task identifies which input patterns activate a neuron. 
One family of solutions searches for image patches within the dataset that maximize the activations of neurons \cite{zhou2014object,zeiler2014visualizing,bau2017network}. Another series of works generates images that maximize certain neuron activations \cite{nguyen2016synthesizing,olah2017feature,erhan2009visualizing,simonyan2013deep,mahendran2015understanding}. 

\textbf{Feature attribution:} Here, the problem is to find what features in the input are important for the response of a neuron. The notion of \emph{importance}/\emph{contribution} is grounded in the effect of removal of a feature on the response. The amount of output change after removing the feature is \emph{marginal contribution}, and the average of marginal contributions of a feature in all possible coalitions with other features in the input is the \emph{Shapley value} \cite{shapley1953value}. Due to computational complexity, several works such as integrated gradients (IntGrad) \cite{Sundararajan2017} and DeepSHAP \cite{Lundberg2017} approximate the Shapley value. Recently it has been shown that many approximations break the axioms \cite{sundararajan2020many}, leaving integrated gradients as a promising candidate.
%

A principal class of feature attribution methods use network gradients.
\cite{simonyan2013deep,baehrens2010explain} propose the input gradient itself as attribution. Guided backpropagation (GBP)~\cite{Springenberg2015}, LRP~\cite{Montavon2017}, and DeepLIFT~\cite{Shrikumar2017} modify gradients during back-propagation. 
Class Activation Maps (CAM)~\cite{zhou2016learning} and GradCAM~\cite{selvaraju2017grad} perform a weighted sum of the last convolutional feature maps. 
Grounded in marginal contribution, 
other approaches (\textit{perturbation methods}) mask the input \cite{fong2017interpretable,fong2019understanding,qi2019visualizing,wagner2019interpretable} or neurons \cite{fong2019understanding,Schulz2020Restricting} and observe the output (or information flow~\cite{Schulz2020Restricting}).

\textbf{Evaluation of feature attribution methods:}  
Early evaluation of interpretability relied on human perception, \eg evaluation by localization accuracy \cite{zhou2016learning} or pointing game \cite{zhang2018top}. However, the model could be using features outside the human annotation or even non-robust features as in \cite{ilyas2019adversarial}, and such localization-based evaluations penalize the correct attribution method. Moreover,~\cite{nie2018theoretical,sixt2019explanations,khakzar2020rethinking,khakzar2019explaining,Adebayo2018} show/prove that several methods with human interpretable attributions generate the same attribution even after the network's weights are randomized. These methods are GBP~\cite{Springenberg2015}, Deconvolution~\cite{zeiler2014visualizing}, DeepTaylor(=LRP-$\alpha1\beta0$)\cite{Montavon2017} and Excitation BackProp \cite{zhang2018top}. 
To evaluate such sensitivity to model parameter randomization, sanity checks \cite{Adebayo2018} have been proposed. Recently, input degradation \cite{Samek2017} and ROAR \cite{hooker2019benchmark} experiments have been introduced for evaluating feature importance. Each of these evaluations measure a different perspective which we explain in section \ref{sec:feature_attribution}.


\section{Selection of critical pathways}

\subsection{Setup and notation}\label{sec:notation}

Consider a neural network $\Phi_{\Theta}(\mathbf{x}):\mathbb{R}^{D} \rightarrow \mathbb{R}$ with $\relu$ activation functions, parameters $\Theta = \{ \theta^{1},...,\theta^{L} \}$, and $L$ hidden layers with $N_{i}$ neurons in layer $i\in \{ 1, ..., L \}$. The total number of neurons is $N = \sum_{i=1}^{L}N_{i}$. We use $\mathbf{z}^{i}\in \mathbb{R}^{N_{i}}$ to represent pre-activation vector at layer $i$, and $\mathbf{a}^{i}\in \mathbb{R}^{N_{i}}$ for representing the corresponding activation vector, where $\mathbf{a}^{i}=\relu(\mathbf{z}^{i})$, $\mathbf{z}^{i}=\theta^{i}\mathbf{a}^{i-1} + \mathbf{b}^{i}$, and $\mathbf{a}^{0}=\mathbf{x}$. Note, our definition of $\Phi_{\theta}(\mathbf{x})$ has a single real valued output. The reason is that we are considering the neural pathway to one neuron, and this neuron could in fact be a hidden neuron in a larger network. 
Thus the response is defined by $\Phi_{\theta}(\mathbf{x}) = \theta^{L+1}\mathbf{a}^{L} + \mathbf{b}^{L+1}$. 
Each individual neuron in layer $i$ is specified by index $j \in \{1,...,N_{i}\}$, thus denoted by $\mathbf{z}^{i}_{j}$ and $\mathbf{a}^{i}_{j}$. The vectors $\mathbf{z}^{i}$ and $\mathbf{a}^{i}$ are specifically associated with input $\mathbf{x}$. The vector containing activations of all $N$ neurons is denoted by $\mathbf{a}=[\mathbf{a}^{i}_{j}]^N$ (same notation for vectors of other entities related to neurons). $\{\mathbf{a}_{j}^{i}\}_{j=1}^{N_{i}}$ denotes the set of neurons in layer $N_{i}$, and $\{0,1\}^{N}$ denotes a set of size $N$ containing $0$s and $1$s.

\subsection{Selection by pruning objective}\label{selection by pruning}

We first describe the pruning objective \cite{lee2018snip} and discuss how \emph{a solution satisfying this objective does not necessitate it being critical}. Let $\mathbf{m}=\{0,1\}^{N}$ be a mask that represents the neurons to be kept and pruned. The pruning objective given an input $\mathbf{x}$ is then defined as:
\begin{equation}\label{eq:pruning_obj}
\begin{split}
\underset{\mathbf{m}}{\arg\min}\mathcal{L}\left(\Phi_{\theta}(\mathbf{x}), \Phi_{\theta}(\mathbf{x}; \mathbf{m}\odot \mathbf{a})\right) \; \; \; \;
\text{s.t.} \; \left \| \mathbf{m} \right \|_{0}\leq \kappa,
\end{split}
\end{equation}
where $\odot$ and $\mathcal{L}$ denote the Hadamard product and the loss respectively. $\kappa$ controls the sparsity.
\Cref{eq:pruning_obj} is a combinatorial optimization problem and a plethora of solutions exist ~\cite{lecun1990optimal, hassibi1993second, han2015learning, molchanov2016pruning, lee2018snip}.

\textbf{Does the pruning objective result in sparse pathways that encode the input?} 
Rectified neural networks have sparse positive activations ~\cite{Glorot2011,olshausen2004sparse,papyan2017convolutional,sulam2018multilayer}, and encode the input in a sparse set of \emph{active} neurons~\cite{olshausen2004sparse,Glorot2011}.
Thus, the network represents discriminative and infrequent features (i.e. features with high information) by sparse/infrequent activation values.
In this regime, positive activations are detectors of features~\cite{bau2017network,erhan2009visualizing,zhou2014object,olah2017feature,zeiler2014visualizing} and encode features that \emph{exist} in the input. Zero activations represent missingness of features. Therefore, if a neuron is dead (zero), its corresponding features do not exist in the input. We also posit that a dead neuron does not contribute to the output (considering zero activation as a baseline for missingness):

\begin{lemma}[Dead Neurons]\label{lemma:deadRelu}
Considering $\mathbf{a}^{i}$ as the input at layer $i$ to the following layers of the network defined by function $\Phi_{\theta}^{>i}(.):\mathbb{R}^{N_{i}} \rightarrow \mathbb{R}$, the Shapley value of a neuron $\mathbf{a}^{i}_{j}$ defined by
$ \sum_{C\subseteq \{\mathbf{a}_{j}^{i}\}_{j=1}^{N_{i}}\setminus \mathbf{a}_{j}^{i} }^{} \frac{|C|!(N_{i}-|C|-1)!}{N_{i}}(\Phi_{\theta}^{>i}(C\cup \mathbf{a}_{j}^{i})- \Phi_{\theta}^{>i}(C))$ is zero if the neuron is dead ($\mathbf{a}^{i}_{j}=0$).
\end{lemma}
\begin{algorithm}[b]\label{alg:greedy}
\SetAlgoLined
 initialize $\mathbf{m}_{j}^{i}=1 \; \; \; \; \forall \: i,j$ \\
 \While{$\left \| \mathbf{m} \right \|_{0}\geq \kappa$}{
  $s_{j}^{i} \leftarrow |\mathbf{a}_{j}^{i}\nabla_{\mathbf{a}_{j}^{i}} \Phi_{\theta}(\mathbf{x})|$ \\
  \lIf{$s_{j}^{i} \leq s_{\kappa} \wedge s_{j}^{i} \neq 0$}{
  $\mathbf{m}_{j}^{i} \leftarrow 0$
  }
 }
 \caption{Pathological greedy pruning}
\end{algorithm}
Lemma \ref{lemma:deadRelu} (proofs for lemmas and propositions are provided in the appendix) shows that dead neurons have a Shapley value of zero, thus do not contribute to the response. We proceed to explain how the pruning objective can select originally dead neurons as critical neurons.
Removing an active neuron results in a change in inputs to next layer's neurons and thus \emph{may} change their activation value, and this \emph{can} result in activating an originally dead neuron. We consider a sparse selected pathway \emph{undesirable} if it contains neurons that were originally dead, but have become active due to the pruning of other neurons. Such a selected pathway is an artificial construct, which does not reflect the original sparse encoding of the input.
\begin{figure}[!t]
    \centering
    \includegraphics[width=\linewidth]{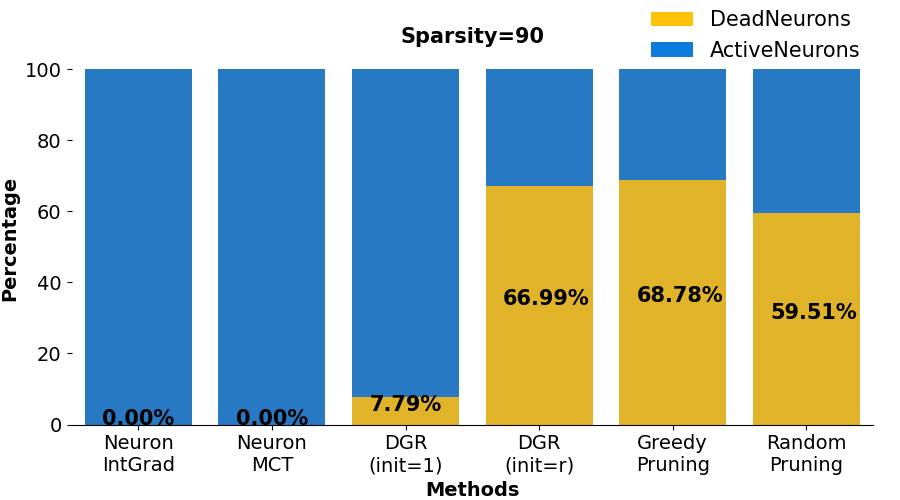}
    \caption{\textbf{Dead Neuron Selection of Pruning Objective.} The percentage of originally dead neurons in the selected pathways of different methods reported for sparsity of 90\% (see appendix for more sparsity values). Evaluation on pathways extracted from VGG-16 on 1k ImageNet images. All pathways selected by pruning objective contain originally dead (now active) neurons. 
    The observation that when selecting the top 10\% of critical neurons, pruning methods select neurons from the dead regions of the network (which after pruning become active) points to the fact that they are selecting pathways unrelated to the input. Our proposal is to use neuron contributions (our NeuronIntGrad, NeuronMCT).}
    \label{fig:paths} 
\end{figure}
\begin{figure*}[!h]
    \centering
    \includegraphics[width=\linewidth]{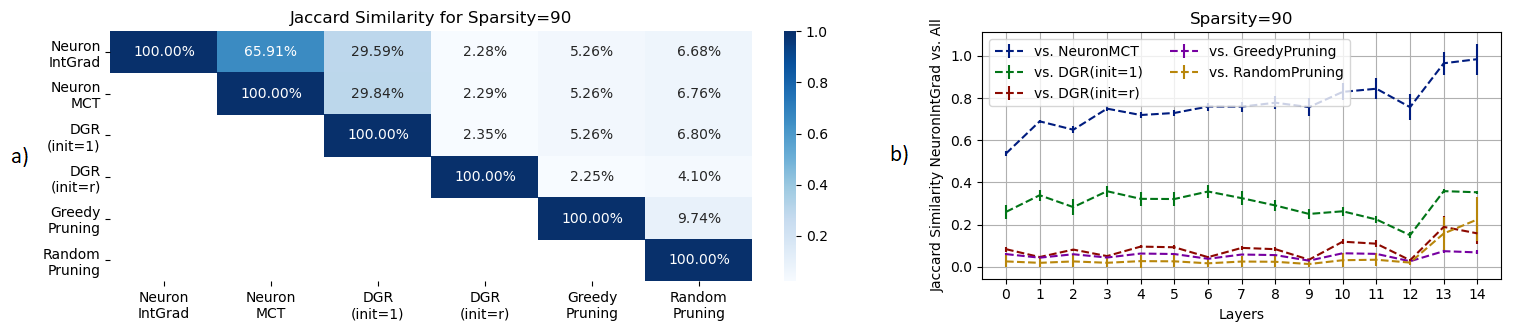}
    \caption{\textbf{Pathway Analysis.} Overlap between pathways of methods: a) in entire network. b) layer-wise overlap between pathways of all methods and NeuronIntGrad. Among the pruning-based methods, only DGR(init=1) does not diverge far from originally active pathways.}
    \label{fig:pathsjaccard}
\end{figure*}

\textbf{Pathological greedy pruning:}
We construct a greedy algorithm that by design searches for irrelevant pathways while solving the pruning objective (\cref{eq:pruning_obj}). The algorithm illustrates how originally dead neurons turn active and become part of the highly sparse selected pathway that produces the same network response.
Our greedy approach (Alg. \ref{alg:greedy}) first ranks all neurons based on their relevance score for the response. The relevance is determined by the effect of removing a neuron, approximated by Taylor expansion similar to \cite{molchanov2016pruning,mozer1989skeletonization}. The relevance score $s_{j}^{i}$ is then:
\begin{equation}\label{eq:greedy_relevance}
s_{j}^{i} = |\Phi_{\theta}(\mathbf{x}) - \Phi_{\theta}(\mathbf{x}; \mathbf{a}_{j}^{i}\leftarrow0)| = |\mathbf{a}_{j}^{i}\nabla_{\mathbf{a}_{j}^{i}} \Phi_{\theta}(\mathbf{x})| \; .
\end{equation}
%
%
%
Next we remove the neuron(s) with the lowest rank, and alternate between rank computation and removal. However, \emph{we tweak the algorithm to find pathways that contain originally dead neurons}.
At each removal step, we remove the lowest contributing neuron that is not dead (without this \emph{crucial step}, dead neurons will be pruned before others as their relevance score is zero). 
By removing a non-zero neuron, the activation pattern can change and some originally dead neurons can activate and become included in the pathway.
Our algorithm illustrates that solving for the pruning objective can result in undesirable pathways. 
%
%
%

\textbf{Distillation Guided Routing}:
DGR~\cite{wang2018interpret} relaxes the pruning objective (\cref{eq:pruning_obj}) by replacing $\mathbf{m}=\{0,1\}^{N}$ with continuous valued gates $0 \leq {\lambda}^{i}_{j} \in \mathbb{R} $. 
To induce sparsity, the objective is regularized with an $\ell_{1}$ norm, \ie $\left \| {\lambda}^{i}_{j} \right \|_{1}$: 
%
\begin{equation}\label{eq:dgr}
\begin{split}
\min_\Lambda{ \mathcal{L}(\Phi_{\theta}(\mathbf{x}) , \Phi_{\theta}(\mathbf{x}; \Lambda \odot \mathbf{a})) + \gamma \sum_{k=1}^{N}\left \| {\lambda}^{i}_{j} \right \|_{1} } \; \; \; \;
\text{s.t.} \; {\lambda}^{i}_{j}\geq0 \; ,
\end{split}
\end{equation}
where $\Lambda=[{\lambda}^{i}_{j}]^{N}$ is the vector of all ${\lambda}^{i}_{j}$.
In our experiments we find that the initial value of $\Lambda$ plays a significant role. Wang \etal \cite{wang2018interpret} use ${\lambda}^{i}_{j}=1 \; \forall i,j$ without discussing its role. We denote different initilizations with DGR(init=\textit{value}).

%
%
%
%
%
%
\subsection{Selection by neuron contribution}\label{select by contrib}
Individual neuron ablation ~\cite{Zhou2018} and network dissection \cite{bau2017network} reveal that specific neurons are critical for certain classes. It is therefore intuitive that critical pathways contain important neurons. The effect of removing a unit, $|\Phi_{\theta}(\mathbf{x}) - \Phi_{\theta}(\mathbf{x}; \mathbf{a}_{j}^{i}\leftarrow0)|$, is called the marginal contribution.
Computing the exact value for the marginal contribution of all neurons is computationally expensive. As we have to ablate each neuron (total $N$) in the network and observe its effect after inference. Therefore we use a Taylor approximation similar to \cref{eq:greedy_relevance}, $\mathbf{c}_{j}^{i} = |\Phi_{\theta}(\mathbf{x}) - \Phi_{\theta}(\mathbf{x}; \mathbf{a}_{j}^{i}\leftarrow0)| = |\mathbf{a}_{j}^{i}\nabla_{\mathbf{a}_{j}^{i}} \Phi_{\theta}(\mathbf{x})|$,
where $\mathbf{c}^{i}_{j}$ denotes the contribution of neuron $\mathbf{a}^{i}_{j}$. 
Pathways selected by this method are hereon referred to as NeuronMCT, where MCT stands for Marginal Contribution Taylor. 
%

The Shapley value is the unique definition that satisfies desirable axioms of feature attribution \cite{Lundberg2017}.
It is defined as the average of marginal contributions of a feature in all possible coalitions with other features in the input. For each neuron, its coalitions with neurons of the same layer are considered. This results in $2^{N_{i}-1}$ possible coalitions. Considering all layers, the total required inference steps becomes $\sum_{i=1}^{L}2^{N_{i}-1}$, which is computationally expensive. Thus we use an approximation method.
The IntGrad \cite{Sundararajan2017} method with baseline 0 is equivalent to the Aumann-Shapley value, which is an extension of the Shapley value to continuous setting \cite{sundararajan2020many}. 

The contribution $\mathbf{c}^{i}_{j}$ using IntGrad with baseline 0 is:
\begin{equation}
    \mathbf{c}^{i}_{j} = \mathbf{a}^{i}_{j}\int_{\alpha = 0}^{1} \frac{\partial \Phi_{\theta} (\alpha \mathbf{a}^{i}_{j}; \mathbf{x}) }{\partial \mathbf{a}^{i}_{j}} \mathrm{d}\alpha
\end{equation}
Henceforth, the contributions assigned as such are referred to as NeuronIntGrad.

\begin{remark}\label{remark:mct_intgrad_assign_zero}
NeuronMCT and NeuronIntGrad assign $\mathbf{c}^{i}_{j}=0$ to a neuron with $\mathbf{a}^{i}_{j}=0$ (dead neuron).
\end{remark}

We denote a pathway by $\mathbf{e}=[\mathbf{e}^{i}_{j}]^N$, where $\mathbf{e}^{i}_{j} \in \{0,1\}$ are indicator variables for each neuron indicating whether neuron belongs to pathway $\mathbf{e}$. Having computed the contributions $\mathbf{c}^{i}_{j}$, in order to select a pathway $\mathbf{e}=[\mathbf{e}^{i}_{j}]^N$, with sparsity value $\kappa$, we select neurons with $\mathbf{c}^{i}_{j} \geq \mathbf{c}_{\kappa}$ where $\mathbf{c}_{\kappa}$ is the contribution value of the corresponding sparsity $\kappa$ in a sorted list of contributions, \ie if $\mathbf{c}^{i}_{j} \geq \mathbf{c}_{\kappa}$ then $\mathbf{e}^{i}_{j}=1$, else $\mathbf{e}^{i}_{j} = 0$. 

Selecting the values higher than $\mathbf{c}_{\kappa}$ is average-case $\mathcal{O}(n)$. Thus the computational burden of selecting a pathway depends on the contribution assignment procedure. NeuronMCT requires one inference, and for NeuronIntGrad we use 50 inference steps in the experiments.
For both methods, the ranking of the $\mathbf{a}_{j}^{i}$ are performed network-wise, and not layer-wise. We directly compare the contribution of neurons from different layers. This is possible because integrated gradients satisfies completeness, and for each layer $i$, $\sum_{j=1}^{N_{i}}\mathbf{c}_{j}^{i} = \Phi_{\theta}(\mathbf{x}) - \Phi_{\theta}(\mathbf{a}^{i}\leftarrow0)$, making the scores directly comparable.
For marginal contribution, by definition the contribution is the change in the output, so the contributions are inherently comparable.
\begin{figure*}[t]
    \centering
    \includegraphics[width=0.99\linewidth]{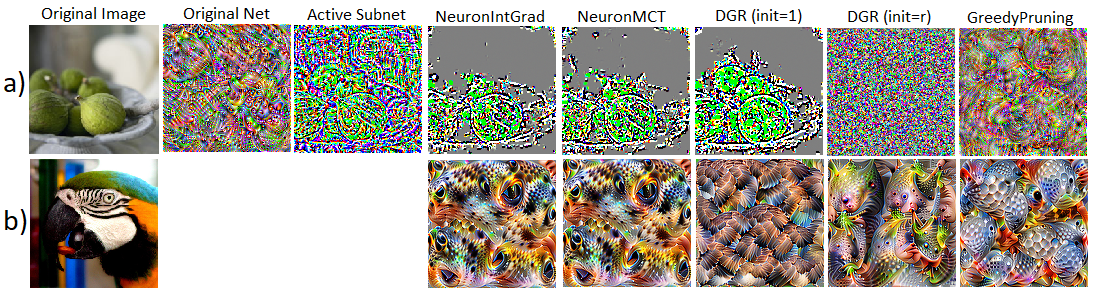}
    \caption{\textbf{Pathway Decoding.} \textbf{a)} Generating an input that maximizes the network response while restricting the network to a specific pathway selected by different methods. For the original network, the generated input contains class-specific (related to "fig" class) features. When restricted to active neurons (Active Subnet), an input similar to the original input is reconstructed. The reconstructed input for our contribution-based methods (NeuronIntGrad, NeuronMCT) contains only the critical input features (the figs) of the original image. The reconstructed images for pruning-based pathways (except DGR(init=1)) do not show any input related information. 
    \textbf{b)} Feature visualization of the top selected neuron of the final convolutional layer in each pathway (this experiment is only relevant for pathway selection methods). The top neuron in NeuronIntGrad and NeuronMCT pathways encodes features related to the bird's eye, which is highly relevant to the input image. The top neurons in pruning-based pathways are associated with features not related to the input. 
    }
    \label{fig:decoding}
\end{figure*}
\begin{figure*}[!t]
    \centering
    \includegraphics[width=0.95\linewidth]{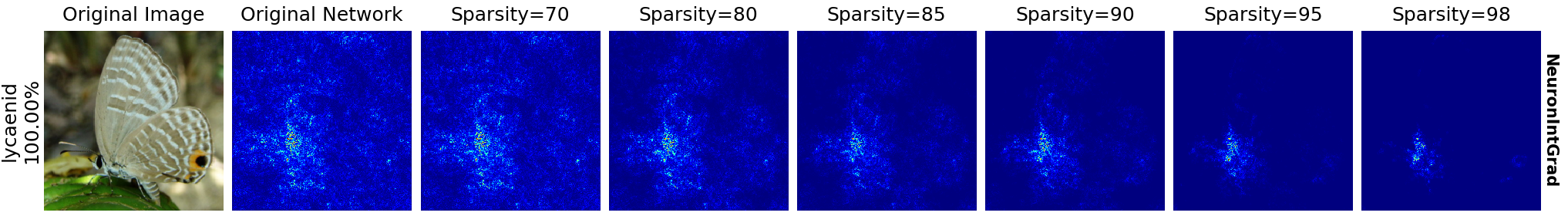}
    \caption{\textbf{Feature Attribution via Pathway Gradient.} The gradients of the locally linear critical pathways at different sparsity levels. The pathway is selected using NeuronIntGrad. The pathways are locally linear and their gradient reflects the critical input features. As we select sparser critical pathways, feature attribution reveals input features that are more critical. More examples in the supplementary.}
    \label{fig:sparsity_grad}
\end{figure*}
\subsection{Pathway selection experiments}\label{sec:path experiments}
\paragraph{Pathway analysis}
To corroborate the claim that the pruning objective results in undesirable pathways, we evaluate the pathways extracted by the discussed methods from a VGG-16 \cite{simonyan2014very} network for 1k ImageNet \cite{deng2009imagenet} images. We report results for the pathways of 90\% sparsity (more values in appendix). \cref{fig:paths} shows the percentage of previously dead neurons in the selected pathways of all methods. We observe that all pruning based methods converge to selecting undesirable pathways. $\sim$69\% of neurons in the top 10\% selected neurons of GreedyPruning are originally dead neurons. Another noteworthy observation is the effect of the initial value of gates in the DGR method. When gates are initialized to 1, the pathways do not drift away from the original active pathway as much as they do with random (uniform $[0, 1]$) initialization (DGR(init=r)). Nevertheless, still $\sim$8\% of neurons in the top 10\% of DGR(init=1) are originally dead neurons. 
We analyze the overlap of the selected pathways using the Jaccard similarity between pathway indicators $\mathbf{e}$ in \cref{fig:pathsjaccard}a. Note the similarity between NeuronIntGrad and DGR(init=1) compared to DGR(init=r). This suggests that when initializing DGR with 1, the selected pathways do not drift significantly to undesired pathways, and they still roughly contain the critical neurons, explaining why \cite{wang2018interpret} observed meaningful pathways. We also perform a layer-wise similarity analysis between pathways in \cref{fig:pathsjaccard}b. We observe that the overlap between pathways of NeuronIntGrad and NeuronMCT increases as we move towards final layers, implying that the overall difference between their pathways is due to differences in earlier layers.

\paragraph*{Pathway decoding}
Feature visualization estimates the input that maximizes a neuron's response.
In order to generate an image $\mathbf{x}_{G}$ that maximizes the response, the network's weights are frozen and optimization by gradient descent is done on the input, \ie ${\arg\max}_{\mathbf{x}_{G}}\Phi_{\theta}(\mathbf{x}_{G})$.
Such optimization without any regularization or priors is prone to generating adversarial artifacts \cite{goodfellow2014explaining,olah2017feature}. Hence, we optimize with preconditioning and transformation robustness techniques as in \cite{olah2017feature} to generated natural looking images. 
The question we are interested in here is what the pathway corresponding to the input can tell us about that input. This allows us to semantically evaluate the pathways derived from different pathway selection methods.
In \cref{fig:decoding}a, we generate inputs that maximize the network response while the network is restricted to different pathways. When considering the original network, features related to the predicted ("Fig") class are visualized. When we restrict the network to the active pathway (Active Subnet), optimization attempts to reconstruct the image.
At 98\% sparsity we observe that, critical features relevant to the predicted class are reconstructed for contribution-based methods. This signifies that the selected sparse pathway has indeed encoded features relevant to the prediction. However, for pathways selected by the pruning objective (except DGR(init=1)), the reconstructions resemble noise.
In \cref{fig:decoding}b, we perform feature visualization (using entire network) of the selected \emph{top neuron} in the final convolutional layer of each pathway. We observe that the selected top neuron by NeuronMCT and NeuronIntGrad is semantically highly relevant to the input, as the neuron is responsible for the bird's eye. The top neuron of DGR(init=1) is also relevant as it relates to feathers. 
While for the other methods, the top selected neuron is semantically irrelevant, further confirming that the selected pathways are not encoding the input.

\section{Interpreting response via critical pathways}
In \cref{sec:linearity} we show that sparse pathways selected by NeuronIntGrad and NeuronMCT are locally linear.  The local linearity is later used in \cref{sec:feature_attribution} for input feature attribution via "pathway gradients" and understanding to which features in the input the pathways correspond.
\subsection{Local linearity of pathways of critical neurons}\label{sec:linearity}
\begin{figure*}[t]
    \centering
    \includegraphics[width=\linewidth]{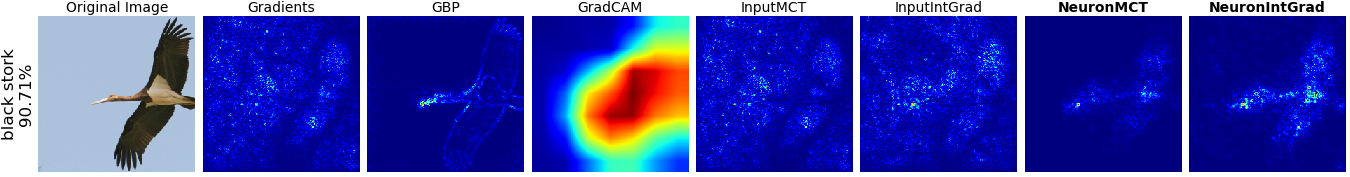}
    \caption{\textbf{Comparison with Attribution Methods.} Results for our method "pathway gradient" are shown on the right (for pathways selected by NeuronMCT and NeuronIntGrad). Our method provides pixel-level explanations as opposed to GradCAM \cite{selvaraju2017grad}. GBP \cite{Springenberg2015} is visually pleasing, but it is merely reconstructing image (Sec. \ref{sec:related_work}). Note the improvement of IntGrad (integrated gradients \cite{Sundararajan2017}) on the neurons (NeuronIntGrad) over IntGrad on input (InputIntGrad). Also note the improvement of marginal contribution on neurons (NeuronMCT) over direct implementation on input (InputMCT = input$\times$gradient \cite{Shrikumar2017}).
    More examples for VGG-16 / ResNet-50 in the appendix.}
    \label{fig:attribution}
\end{figure*}
\begin{figure*}[h]
    \centering
    \includegraphics[width=0.9\linewidth]{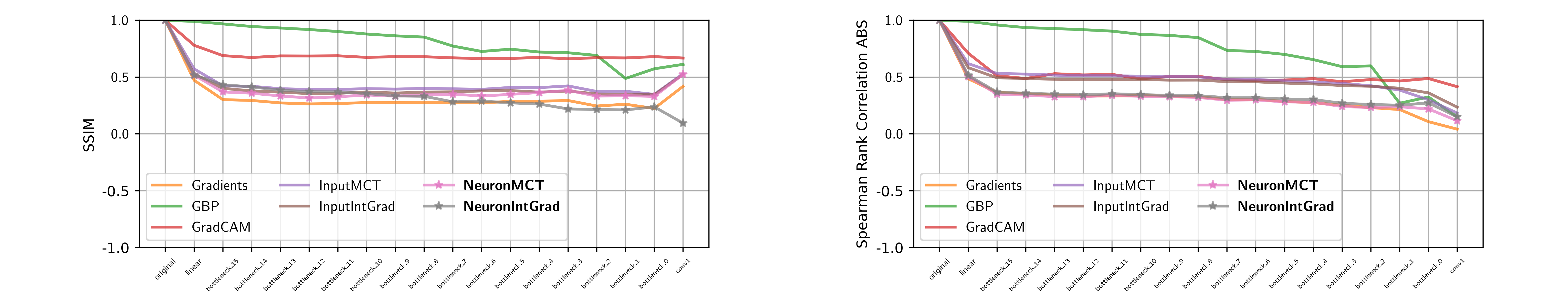}
    \caption{\textbf{Randomization-Sensitivity Sanity Check. \cite{Adebayo2018}} Similarity of attributions before and after network (ResNet-50, ImageNet) parameter randomization. High similarity after randomization suggests that the attribution method is not explaining the network.}
    \label{fig:sanity_vis}
\end{figure*}
Networks with piecewise linear activation functions are piecewise linear in their output domain \cite{Montufar2014}, and thus are linear at a specific point \textbf{x}, and $\forall i,j$:
\begin{equation}
    \Phi_{\theta}(\mathbf{x}) = (\nabla_{\mathbf{x}}\Phi_{\theta}(\mathbf{x}))^{\top }\mathbf{x} + \textbf{b}^{L+1} \; ; \; \mathbf{z}^{i}_{j} = (\nabla_{\mathbf{x}}\mathbf{z}^{i}_{j})^{\top }\mathbf{x} + \mathbf{b}^{i}_{j}
\end{equation}
Although the network degenrates into a linear function at a given point, \emph{it does not mean it is locally linear}. Indeed both the value \cite{goodfellow2016deep} and the gradient \cite{ghorbani2019interpretation} are unstable around a point. To discuss the local linearity of the rectified network, we need to define activation pattern~\cite{Raghu2017,Lee2019towardsLinear}:
\begin{definition}[Activation Pattern($\mathcal{AP}$)] $\mathcal{AP}$ is a set of indicators for neurons denoted by $\mathcal{AP}=\{\mathbbm{1}(\mathbf{a}^{i}_{j})\}^{N}$ where $\mathbbm{1}(\mathbf{a}^{i}_{j})=1$ if $\mathbf{a}^{i}_{j}>0$ and $\mathbbm{1}(\mathbf{a}^{i}_{j})=0$ if $\mathbf{a}^{i}_{j}\leq 0$.
\end{definition}
The feasible set $S(\mathbf{x})$ of an $\mathcal{AP}$ is the input regions where the $\mathcal{AP}$ is constant and thus the function is linear.
Let $\mathcal{B}(\mathbf{x})_{\epsilon,2} = \{\bar{\mathbf{x}} \in \mathbb{R}^{D}: ||\bar{\mathbf{x}}-\mathbf{x}||_{2} \leq \epsilon \}$ denote the  $\ell_{2}$-ball around $\mathbf{x}$ with radius $\epsilon$, and let $\hat{\epsilon}_{\mathbf{x},2}$ denote the largest $\ell_{2}$-ball around $\mathbf{x}$ where the $\mathcal{AP}$ is fixed, \ie
\begin{equation}
\hat{\epsilon}_{\mathbf{x},2} \doteq \underset{\epsilon \geq 0:\mathcal{B}_{\epsilon,2}(\mathbf{x})\subseteq S(\mathbf{x})}{\max}\epsilon
\end{equation}
%
%
%
$\hat{\epsilon}_{\mathbf{x},2}$ is the minimum $\ell_{2}$ distance between $\mathbf{x}$ and the corresponding hyperplanes of all neurons $\mathbf{z}^{i}_{j}$ \cite{Lee2019towardsLinear}. The hyperplane defined by neuron $\mathbf{z}^{i}_{j}$ at point $\mathbf{x}$ is $\{\bar{\mathbf{x}} \in \mathbb{R}^{D} : (\nabla_{\mathbf{x}}\mathbf{z}^{i}_{j})^{\top }\bar{\mathbf{x}} + \mathbf{b}=0\}$ or $\{\bar{\mathbf{x}} \in \mathbb{R}^{D} : (\nabla_{\mathbf{x}}\mathbf{z}^{i}_{j})^{\top }\bar{\mathbf{x}} + (\mathbf{z}^{i}_{j} - (\nabla_{\mathbf{x}}\mathbf{z}^{i}_{j})^{\top }\mathbf{x})=0\}$. 
If $\nabla_{\mathbf{x}}\mathbf{z}^{i}_{j} \neq 0$ then the distance between $\mathbf{x}$ and $\mathbf{z}^{i}_{j}$ is
\begin{equation}
| (\nabla_{\mathbf{x}}\mathbf{z}^{i}_{j})^{\top }\mathbf{x} + (\mathbf{z}^{i}_{j} - (\nabla_{\mathbf{x}}\mathbf{z}^{i}_{j})^{\top }\mathbf{x})| / ||\nabla_{\mathbf{x}}\mathbf{z}^{i}_{j}||_{2} = |\mathbf{z}^{i}_{j}|/||\nabla_{\mathbf{x}}\mathbf{z}^{i}_{j}||_{2}
\end{equation}
%
%
%
For a neuron $\mathbf{z}^{i}_{j}$, if $\nabla_{\mathbf{x}}\mathbf{z}^{i}_{j}=0$, then $\mathbf{z}^{i}_{j}=\mathbf{b}^{i}_{j}$. In order for the activation of this neuron to change, the $\nabla_{\mathbf{x}}\mathbf{z}^{i}_{j}$ and consequently the $\mathcal{AP}$ has to change. Therefore the distance is goverened by neurons for which $\nabla_{\mathbf{x}}\mathbf{z}^{i}_{j} \neq 0$.
\cite{Lee2019towardsLinear} prove that $\hat{\epsilon}_{\mathbf{x},2}=\underset{i,j}{\min}|\mathbf{z}^{i}_{j}|/||\nabla_{\mathbf{x}}\mathbf{z}^{i}_{j}||_{2}$. Since $\nabla_{\mathbf{x}}\mathbf{z}^{i}_{j} \neq 0$, the existence of a linear region $\hat{\epsilon}_{\mathbf{x},2}$ depends on $|\mathbf{z}^{i}_{j}|$ not being zero.

\textbf{Locally linear network approximation}: In order to approximate the original model $\Phi_{\theta}(\mathbf{x})$ with a selected pathway $\mathbf{e}$, we replace each neuron $\mathbf{a}^{i}_{j}$ which is not in the pathway, \ie $\mathbf{e}^{i}_{j}=0$ with a constant value equal to the initial value ($\mathbf{a}^{i}_{j}$) of that neuron. Note the new constant is not a neuron anymore and thus does not propagate gradient. Replacing the neuron with its initial value keeps $\mathcal{AP}$, and neurons $\mathbf{z}^{i}_{j}$ unchanged. We denote such an approximate model by $\hat{\Phi}_{\theta}(\mathbf{x};\mathbf{e})$. Proofs are provided in the appendix.

\begin{proposition}\label{prop:linearity}
In a $\relu$ neural network $\Phi_{\theta}(\mathbf{x}):\mathbb{R}^{D} \rightarrow \mathbb{R}$, for a pathway defined by $[\mathbf{e}^{i}_{j}]^{N}$, if $\mathbf{a}^{i}_{j} > 0 \; \forall \; \mathbf{e}^{i}_{j}=1$, then there exists a linear region $\hat{\epsilon}_{\mathbf{x},2}>0$ for $\hat{\Phi}_{\theta}(\mathbf{x};\mathbf{e})$ at $\mathbf{x}$.
\end{proposition}
\begin{proposition}\label{prop:mct_intgrad_linear}
Using NeuronIntGrad and NeuronMCT, if $\mathbf{c}_{\kappa}>0$, then $\hat{\Phi}_{\theta}(\mathbf{x};\mathbf{e})$ at $\mathbf{x}$ is locally linear.
\end{proposition}
%
%
%
%

\subsection{Input feature attribution via critical pathways}\label{sec:feature_attribution}
The gradient of a linear model represents the contributions of each corresponding input feature\cite{mishra2017local,baehrens2010explain,simonyan2013deep}. Therefore several works perform a first-order Taylor approximation of the network \cite{baehrens2010explain,simonyan2013deep}. However, the gradients of networks are unstable and drastically change around an input. To capture the true direction of change,
SmoothGrad \cite{smilkov2017smoothgrad} averages gradients and LIME \cite{mishra2017local} fits a linear model to the input neighborhood. However, for a locally linear network, the gradients are already stable (constant within a region) in the linear region neighborhood. The gradient reflects the contributions of features in that neighborhood.
Based on Proposition~\ref{prop:mct_intgrad_linear}, the approximate model $\hat{\Phi}_{\theta}(\mathbf{x};\mathbf{e})$ is locally linear for NeuronMCT and NeuronIntGrad. Thus, we can derive linear approximations for the model using the critical pathways of the model and use their gradient as attribution maps. We refer to this method as ``pathway gradient''. We use different levels of sparsity and observe the most critical input features for the response (\cref{fig:sparsity_grad}). The attribution methodology is visually compared with other attribution methods in \cref{fig:attribution}. For pathways selected via pruning, as their pathways do not contain critical input information, it would be senseless to use them for feature attribution. There is no guarantee for their linearity as Proposition \ref{prop:linearity} requires all neurons within the pathway to be active.

\textbf{Computing contribution of neurons vs. input pixels:}
Computing the Shapley value [51] for \emph{pixels} does not account for correlations between pixels\footnote{Correlation and interaction are different. The latter is related to effect of features in different coalitions which is accounted for by Shapley value}. 
Ideally, one should know which pixels are correlated (e.g. belong to the same object), and compute a single Shapley value for this group.
The Shapley value for a group is known as the Generalized Shapley value~\cite{marichal2007axiomatic}. 
There is an exponential number of groups of pixels. We thus aim to compute the Shapley value only for groups of \textit{correlated} pixels \eg an object (more in the appendix).
Within the pathway gradient framework, we compute the Shapley value for \emph{neurons} instead. Neurons inherently correspond to correlated groups of pixels. Thus we are indirectly computing the Shapley value (contribution) of those correlated groups of pixels.
In our experiments, using MCT and IntGrad on neurons (denoted by NeuronMCT and NeuronIntGrad) results in considerably better attributions compared to applying them only to input pixels (denoted by InputMCT and InputIntGrad).

\textbf{Baseline choice}: The baseline in feature attribution represents the absence of a feature. In the image domain, \cite{zeiler2014visualizing,Sundararajan2017} consider the zero (black image) as baseline. However, zero pixel values do not necessarily reflect the absence of a feature \cite{sturmfels2020visualizing,izzo2020baseline}. As explained in Sec.~\ref{selection by pruning}, zero neurons represent missingness in sparse rectifier networks. The zero baseline is therefore, more justified for neurons than for the input space, and has been also used in \cite{Ancona2019,Shrikumar2017}. 

\begin{figure*}[!h]
    \centering
    \includegraphics[width=\linewidth]{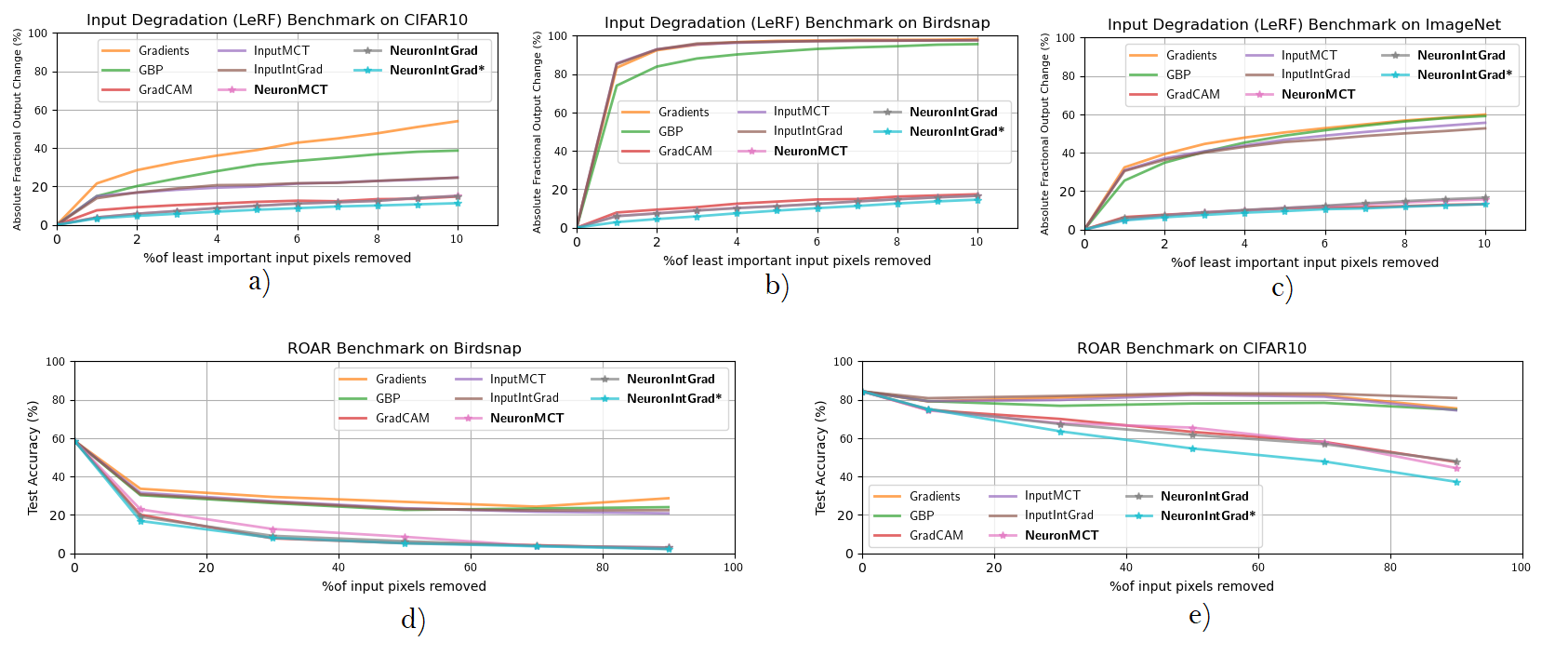}
    \caption{\textbf{Feature Importance.} a,b,c) \textbf{Input degradation-LeRF}: Removes input pixels based on their importance (least relevant features first) and measures output change. Represents how well the method avoids attributing the response to unimportant features (Bridsnap, and ImageNet). d,e) \textbf{ Remove and retrain (ROAR)}: Removes top 10/30/50/70/90 percent of important pixels and retrains on the modified inputs (Birdsnap, Cifar10). If the accuracy does not drop, the attribution method is not highlighting important features.}
    \label{fig:roar_lerf}
\end{figure*}
\subsection{Feature attribution evaluation experiments}
Grounding attribution in theoretical notions such as Shapley value is desired \cite{Lundberg2017,sundararajan2020many} (GradCAM, GBP, and Gradients are not based on this notion). However, experiments can point to specific shortcomings in methods, e.g. the approximate Shapley value (IntGrad\cite{Sundararajan2017}) for pixels does not perform well in experiments, which may be due to disregarding correlations between input pixels.
Each of our experiments examines the methods from a different perspective. As explained in Sec.~\ref{sec:related_work}, visual evaluation can be unreliable. 
Network parameter randomization sanity checks \cite{Adebayo2018} evaluate whether the method is explaining model behavior. Input degradation \cite{Samek2017} and Remove-and-Retrain (ROAR) \cite{hooker2019benchmark} evaluate whether the attribution maps are showing important features in the input  (refer to appendix for details). We use TorchRay~\cite{fong2019understanding} for implementation of other attribution methods.

%

\textbf{Network parameter randomization sanity checks \cite{Adebayo2018}}: Several attribution methods, such as LRP$-\alpha1\beta0$ \cite{Montavon2017}, Excitation Backprop \cite{zhang2018top}, and (GBP)\cite{Springenberg2015} generate the same result after the network is randomly initialized, thus they are not explaining the network \cite{Adebayo2018,sixt2019explanations}.
In this experiment, parameters of the network are successively replaced by random weights, from the last layer to the first layer. At each step, the similarity between the attributions from the original and randomized network are reported (\cref{fig:sanity_vis}) for 1k ImageNet images (ResNet-50).
It is noteworthy that GradCAM seems sensitive here, but the experiment is unfair to it due to the low dimensionality of its maps \cite{Adebayo2018}. Our attributions confidently pass this sanity check.

\textbf{Input degradation - LeRF \cite{Samek2017}}: Pixels are removed based on their attribution score and the output change is measured. We remove least relevant features first (LeRF).
LeRF evaluates methods based on sufficiency of the features for classification and how well methods avoid assigning scores to unimportant features. Results are reported for ResNet-50 on ImageNet, Birdsnap, and Cifar.
We observe (\cref{fig:roar_lerf}(a,b,c)) considerable improvement of NeuronIntGrad and NeuronMCT over InputIntGrad and InputMCT.

\textbf{Remove and retrain (ROAR) \cite{hooker2019benchmark}}: 
In input degradation experiments, the change in output might be a result of the network not having seen such artifacts during training. Therefore the ROAR benchmark retrains the network on the modified images.
If the accuracy does not drop, the attribution method is not highlighting important features. The experiment is performed for different percentiles of removed pixels. Due to the large number (105) of retraining sessions required, we exclude ImageNet and use Cifar10 (ResNet-8) and Bridsnap (ResNet-50) datasets. Gradients, GBP, InputMCT and InputIntGrad are not revealing features that the model learns to use during training.
We observe that NeuronIntGrad and NeuronMCT immensely improve their input counterparts (\cref{fig:roar_lerf}d,e). NeuronIntGrad and NeuronMCT are performing equally to GradCAM. GradCAM benefits from interpolation and has smooth heatmaps, which seem to help with ROAR performance. We see that smoothing (by morphological opening) on our methods (referred to by * in \cref{fig:roar_lerf}), performs best. 

In summary, the attribution experiments show that attribution via critical pathways is a valid methodology with fine-grained attributions. Fine-grained (pixel-level) attributions convey more accurately which features are important to users (\cref{fig:attribution}). The results support that computing marginal contribution and the Shapley value for neurons improves attribution over directly computing them for input pixels. We posit that (Sec.~\ref{sec:feature_attribution}) this can be due to the Shapley value for pixels not accounting for correlations between them. Whereas, computing the Shapley value of neurons implicitly considers correlations between pixels. 
The feature attribution experiments also validate that selected pathways using neuron contributions indeed correspond to critical input features.

\section{Conclusion}
We demonstrate that solving the pruning objective does not necessarily yield pathways that encode critical input features. We propose finding critical pathways based on the neurons’ contributions to the response, and show how these sparse pathways can be leveraged for interpreting the neural response by proposing the ``pathway gradient'' feature attribution method.
Our findings on critical pathways and pruning imply that we may need to revisit the reliability of pruned networks, and point to the direction of pruning via attribution. Moreover, considering critical pathways can be of value to other interpretation approaches, \eg restricting the network to critical pathways can serve as a possible remedy to the vulnerability of perturbation-based attribution methods to adversarial solutions. 

\section*{Acknowledgments}
The authors acknowledge the support of the Munich Center for Machine Learning (MCML) and partial support of Siemens Healthineers. C. Rupprecht is supported by Innovate UK (project 71653) on behalf of UK Research and Innovation (UKRI) and by the European Research Council (ERC) IDIU-638009. A. Khakzar and S.T. Kim are corresponding authors.

{\small
\bibliographystyle{ieee_fullname}
\bibliography{egbib}
}

\appendix
\section{Proofs}\label{app:proof}

\subsection{Proof of Lemma \ref{lemma:deadRelu}}\label{pf:deadRelu}

\addtocounter{definition}{-5}
\begin{lemma}[Dead Neurons]
Considering $\mathbf{a}^{i}$ as the input at layer $i$ to the following layers of the network defined by function $\Phi_{\theta}^{>i}(.):\mathbb{R}^{N_{i}} \rightarrow \mathbb{R}$, the Shapley value of a neuron $\mathbf{a}^{i}_{j}$ defined by
$ \sum_{C\subseteq \{\mathbf{a}_{j}^{i}\}_{j=1}^{N_{i}}\setminus \mathbf{a}_{j}^{i} }^{} \frac{|C|!(N_{i}-|C|-1)!}{N_{i}}(\Phi_{\theta}^{>i}(C\cup \mathbf{a}_{j}^{i})- \Phi_{\theta}^{>i}(C))$ is zero if the neuron is dead ($\mathbf{a}^{i}_{j}=0$).
\end{lemma}

For any layer $i$, the Shapley value (with baseline zero) of a neuron $\mathbf{a}_{j}^{i}$ is defined as:
\begin{equation}
    \sum_{C\subseteq \{\mathbf{a}_{j}^{i}\}_{j=1}^{N_{i}}\setminus \mathbf{a}_{j}^{i} }^{} \frac{|C|!(N_{i}-|C|-1)!}{N_{i}}(\Phi_{\theta}^{>i}(C\cup \mathbf{a}_{j}^{i})- \Phi_{\theta}^{>i}(C)) \; ,
\end{equation}

where $\Phi_{\theta}^{>i}$ denotes the neural function after layer $i$. The input to $\Phi_{\theta}^{>i}$ is the activation vector $\mathbf{a}^{i}$.  We need to show that for all $\mathbf{a}_{j}^{i}$ and all possible coalitions $C\subseteq \{\mathbf{a}_{j}^{i}\}_{j=1}^{N_{i}}\setminus \mathbf{a}_{j}^{i}$:
\begin{equation}
    \Phi_{\theta}(C\cup \mathbf{a}_{j}^{i} ;\mathbf{x}) = \Phi_{\theta}(C;\mathbf{x}) \; .
\end{equation}

We know for any $\mathbf{a}^{i}$ the outputs of neurons in the next layer are:
\begin{equation}
    \mathbf{z}^{i+1}=\theta^{i+1}\mathbf{a}^{i} + \mathbf{b}^{i+1} \; .
\end{equation}

As the baseline is considered zero, ablating a neuron $\mathbf{a}_{j}^{i}$ is done by $\mathbf{a}_{j}^{i}=0$. Thus $\mathbf{z}^{i+1}$ does not change by ablation of $\mathbf{a}_{j}^{i}$ for any coalition $C$. As $\mathbf{z}^{i+1}$ does not change, $\Phi_{\theta}$ does not change, thus we get $\Phi_{\theta}(C\cup \mathbf{a}_{j}^{i} ;\mathbf{x}) = \Phi_{\theta}(C;\mathbf{x})$.

\subsection{Proof of Proposition \ref{prop:linearity}}\label{pf:linearity}

\addtocounter{definition}{+2}
\begin{proposition}
In a $\relu$ rectified neural network with $\Phi_{\theta}(\mathbf{x}):\mathbb{R}^{D} \rightarrow \mathbb{R}$, for a path defined by $[\mathbf{e}^{i}_{j}]^{N}$, if $\mathbf{a}^{i}_{j} > 0 \; \forall \; \mathbf{e}^{i}_{j}=1$, then there exists a linear region $\hat{\epsilon}_{\mathbf{x},2} > 0$ for $\hat{\Phi}_{\theta}(\mathbf{x};\mathbf{e})$ at $\mathbf{x}$.
\end{proposition}

The linear region, $\hat{\epsilon}_{\mathbf{x},2}$ is the largest $\ell_{2}$-ball around $\mathbf{x}$ where the $\mathcal{AP}$ is fixed, \ie
\begin{equation}
\hat{\epsilon}_{\mathbf{x},2} \doteq \underset{\epsilon \geq 0:\mathcal{B}_{\epsilon,2}(\mathbf{x})\subseteq S(\mathbf{x})}{\max}\epsilon
\end{equation}
%
%
%
$\hat{\epsilon}_{\mathbf{x},2}$ is the minimum $\ell_{2}$ distance between $\mathbf{x}$ and the corresponding hyperplanes of all neurons $\mathbf{z}^{i}_{j}$ \cite{Lee2019towardsLinear}.
In \cref{sec:linearity} we discuss that the distance is goverened by neurons for which $\nabla_{\mathbf{x}}\mathbf{z}^{i}_{j} \neq 0$. We have
\cite{Lee2019towardsLinear} $\hat{\epsilon}_{\mathbf{x},2}=\underset{i,j}{\min}|\mathbf{z}^{i}_{j}|/||\nabla_{\mathbf{x}}\mathbf{z}^{i}_{j}||_{2}$. Thus, the existence of a linear region $\hat{\epsilon}_{\mathbf{x},2}$ depends on $|\mathbf{z}^{i}_{j}|$ not being equal to zero.
%
We are selecting a path $[\mathbf{e}^{i}_{j}]^{N}$ where for each neuron $\mathbf{a}^{i}_{j} > 0$ and thus we have $\mathbf{z}^{i}_{j} > 0$. If we replace every neuron not on the path with a constant value equivalent to the original value of the activation of that neuron, the activation pattern $\mathcal{AP}$ remains constant, and thus we get a new approximate neural network $\hat{\Phi}_{\theta}(\mathbf{x};\mathbf{e})$ at $\mathbf{x}$, where all neurons $\mathbf{z}^{i}_{j} > 0$. Therefore $\hat{\epsilon}_{\mathbf{x},2} \neq 0$ and there exists a linear region.
\subsection{Proof of Proposition \ref{prop:mct_intgrad_linear}}\label{pf:mct_intgrad_linear}

\begin{proposition}
Using NeuronIntGrad and NeuronMCT, if $\mathbf{c}_{\kappa}>0$, then $\hat{\Phi}_{\theta}(\mathbf{x};\mathbf{e})$ at $\mathbf{x}$ is locally linear.
\end{proposition}

For NeuronMCT and NeuronIntGrad the contributions are assigned by:
\begin{equation}
\mathbf{c}_{j}^{i} = |\Phi_{\theta}(\mathbf{x}) - \Phi_{\theta}(\mathbf{x}; \mathbf{a}_{j}^{i}\leftarrow0)| = |\mathbf{a}_{j}^{i}\nabla_{\mathbf{a}_{j}^{i}} \Phi_{\theta}(\mathbf{x})|
\end{equation}
and
\begin{equation}
    \mathbf{c}^{i}_{j} = \mathbf{a}^{i}_{j}\int_{\alpha = 0}^{1} \frac{\partial \Phi_{\theta} (\alpha \mathbf{a}^{i}_{j}; \mathbf{x}) }{\partial \mathbf{a}^{i}_{j}} \mathrm{d}\alpha
\end{equation}
respectively. It is evident that if $\mathbf{c}_{j}^{i} \neq 0$ then $\mathbf{a}_{j}^{i}\neq0$. Therefore a path selected by  $|\mathbf{c}_{j}^{i}| > 0$ we have all $\mathbf{a}_{j}^{i} > 0$. Hence according to Prop.~\ref{prop:linearity} the selected paths and the approximate $\hat{\Phi}_{\theta}(\mathbf{x};\mathbf{e})$ is locally linear.


\subsection{Proofs for axioms of marginal contribution}

Defining marginal contribution of neuron $\mathbf{a}^{i}_{j}$ at layer $i$ as:
\begin{equation}\label{appendix:eq:marginal}
    \mathbf{c}^{i}_{j} = \Phi_{\theta}^{>i}(\{\mathbf{a}^{i}_{j} \}_{j=1}^{N_{i}}) - \Phi_{\theta}^{>i}(\{\mathbf{a}^{i}_{j} \}_{j=1}^{N_{i}}\setminus\mathbf{a}^{i}_{j})
\end{equation}
\subsubsection{Null player} 

The null player axiom asserts that if a neuron is a null player, \ie

\begin{equation}\label{eq:null_player}
    \Phi_{\theta}^{>i}(S \cup \mathbf{a}^{i}_{j}) = \Phi_{\theta}^{>i}(S) \; ,
\end{equation}

for all $S \subseteq \{\mathbf{a}^{i}_{j} \}_{j=1}^{N_{i}} \setminus \mathbf{a}^{i}_{j}$, then $\mathbf{c}^{i}_{j}$ must be zero.

Eq. \ref{eq:null_player} is assumed for all $S$, therefore by substituting $S = \{\mathbf{a}^{i}_{j} \}_{j=1}^{N_{i}} \setminus \mathbf{a}^{i}_{j}$, in Eq. \ref{eq:null_player} we get:

\begin{equation}
    \Phi_{\theta}^{>i}(\{\mathbf{a}^{i}_{j} \}_{j=1}^{N_{i}}) = \Phi_{\theta}^{>i}(\{\mathbf{a}^{i}_{j} \}_{j=1}^{N_{i}}\setminus\mathbf{a}^{i}_{j}) \; ,
\end{equation}

which results in $\mathbf{c}^{i}_{j}=0$.
\subsubsection{Symmetry} 

The symmetry axiom asserts for all $S \subseteq \{\mathbf{a}^{i}_{j} \}_{j=1}^{N_{i}} \setminus \{\mathbf{a}^{i}_{j}, \mathbf{a}^{i}_{k}\}$ if

\begin{equation}\label{eq:symmetry}
    \Phi_{\theta}^{>i}(S \cup \mathbf{a}^{i}_{j}) = \Phi_{\theta}^{>i}(S \cup \mathbf{a}^{i}_{k})
\end{equation}

holds, then $\mathbf{c}^{i}_{k}=\mathbf{c}^{i}_{j}$. 

Eq. \ref{eq:symmetry} is assumed for all $S$, therefore by substituting $S = \{\mathbf{a}^{i}_{j} \}_{j=1}^{N_{i}} \setminus \{\mathbf{a}^{i}_{j}, \mathbf{a}^{i}_{k}\}$, in Eq. \ref{eq:null_player} we have:
\begin{equation}
    \Phi_{\theta}^{>i}(\{\mathbf{a}^{i}_{j} \}_{j=1}^{N_{i}}\setminus\mathbf{a}^{i}_{j}) = 
\Phi_{\theta}^{>i}(\{\mathbf{a}^{i}_{j} \}_{j=1}^{N_{i}}\setminus\mathbf{a}^{i}_{k}) \; .
\end{equation}

By substituting into Eq. \ref{appendix:eq:marginal}, we obtain $\mathbf{c}^{i}_{k}=\mathbf{c}^{i}_{j}$.

\section{Further Discussions}\label{app:discussion}
\subsection{Computing contribution of neurons vs. pixels}\label{app:discuss_interdependeny}
If we compute the marginal contribution or Shapley value for a single feature of the input, \eg a pixel, the distributional interdependencies, and correlations between the pixels are not considered. This is not to be confused with the interdependency that the Shapley value considers by taking different coalitions into account.
For instance, ablating a single pixel from an object in an image does not affect the score of an Oracle classifier, in any coalition. One must consider that all the pixels are related and exist in one object when computing the marginal contribution and Shapley value for the object (all pixels considered as one feature). Shapley value of a set of features is known as generalized Shapley value ~\cite{marichal2007axiomatic}.
We can observe a consequence of this phenomenon, in the different results obtained by \cite{ancona2017towards} when removing single pixels (occlusion-1) or removing patches, where the latter results in more semantic attribution maps.
Several works implicitly consider such correlations by masking a group of pixels. The question is what mask should we look for, as the prior information about the dependency of pixels is not available. There are $2^{N}$ possible masks that one can select. Moreover, a larger mask containing a feature might get the same or higher contribution score as the mask of the feature. Therefore in \cite{fong2017interpretable,fong2019understanding} priors such as the size of the mask are used. These methods look for the smallest masks with the highest contribution. In the regime of neural networks, we encounter more problems with mask selection. If we do not enforce any prior, we can get adversarial masks \cite{fong2017interpretable,goodfellow2014explaining}. Therefore, several works\cite{fong2017interpretable,fong2019understanding} enforce priors such as smoothness of the masks. 
On the other hand, if we use the prior encoded in the network (which is learned from the distribution of the data), we implicitly consider the group of pixels that are correlated with each other. Thus by computing the contribution of individual neurons, we are considering a complex group of pixels and their distributional relationships.



\section{Implementation details}\label{app:implementation}

The sparsity level of ResNet-50 is 70\% and VGG-16 is 90\% in the experiments, unless stated otherwise.

\subsection{Network parameter randomization sanity check \cite{Adebayo2018}}
All attribution methods are run on ResNet50 (PyTorch pretrained) and on 1k ImageNet images. The acquired attribution maps from all methods are normalized to [-1 1] as stated by \cite{Adebayo2018}. The layers are randomized from a normal distribution with mean=0 and variance=0.01 in a cascading manner from the last to the first. After the randomization of each layer, the similarity metrics (SSIM and Spearman Rank Correlation) are calculated between the map from the new randomized model and the original pretrained network. Methods that are not sensitive to network parameters (like GBP) would hence lead to high levels of similarity between maps from normalized networks and the original map.

\subsection{Input degradation - LeRF \cite{Samek2017}}
We report results on CIFAR using a custom ResNet8 (three residual blocks), Birdsnap using ResNet-50, and ImageNet (validation set) using ResNet-50. We show the absolute fractional change of the output as we remove the least important pixels. Lower curves mean higher specificity of the methods. Note that, for NeuronMCT and NeuronIntGrad, the pixel perturbation process is performed on the original model not on the critical paths selected by these methods. The critical paths are only used to obtain the attribution maps and not after.

\subsection{Remove and Retrain (ROAR) \cite{hooker2019benchmark}}
We perform the experiments with top {30; 50; 70; 90} of pixels perturbed. The model is retrained for each attribution method (8 methods) on each percentile (5 percentiles) 3 times. Due to the large number of retraining sessions required, we cannot report this benchmark on other datasets. We evaluate this benchmark on CIFAR-10 (60k images, 32x32) with a ResNet-8 (three residual blocks).

\clearpage
\section{Supplementary results}\label{app:results}

\begin{table*}[!h]
\caption{\textbf{ROAR}: AUCs reported for each attribution method. The lower the AUC the better.}
\label{table:roar}
\centering
\begin{small}
\begin{tabular}{|l|l|l|l|l||l||l|l|l|}
\hline
 & Gradient & GBP & GradCAM & InputMCT & InputIntGrad & NeuronMCT & NeuronIntGrad & NeuronIntGrad*\\
\hline
Cifar-10    &  0.728  &   0.702   &     0.584    &   0.723    &   0.741   & 0.580   &  0.574 &  \textbf{0.524}\\
\hline
Birdsnap     &  0.269  &    0.243   &    0.096    &   0.242    &    0.242   &  0.117   &  0.099 &  \textbf{0.090}\\
\hline
\end{tabular}
\end{small}
\end{table*}



\begin{table*}[!h]
\caption{\textbf{Input degredation (LeRF)}: AUCs reported for each attribution method. The lower the AUC the better.}
\label{table:LerF}
\centering
\begin{small}
\begin{tabular}{|l|l|l|l|l||l||l|l|l|}
\hline
  & Gradient & GBP & GradCAM & InputMCT & InputIntGrad & NeuronMCT & NeuronIntGrad & NeuronIntGrad*\\
\hline
Cifar-10    &  0.037  &   0.028   &    0.010    &   0.019    &   0.019   & 0.009   &  0.009 & \textbf{0.007}\\
\hline
Birdsnap     &  0.090  &   0.085   &    0.012    &   0.090    &   0.090   & 0.010   &  0.010 &  \textbf{0.008}\\
\hline
ImageNet     &  0.046  &   0.044   &    0.009    &   0.043    &   0.041   & 0.010   &  0.010 &  \textbf{0.009}\\
\hline
\end{tabular}
\end{small}
\end{table*}

\begin{figure}[!h]
    \centering
    \includegraphics[width=0.99\columnwidth]{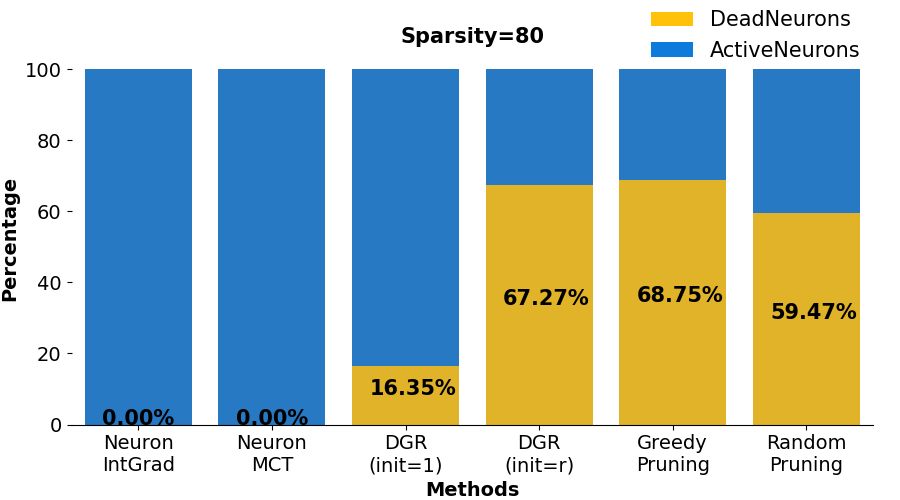}
    \caption{\textbf{Dead Neuron Selection of Pruning Objective (Sparsity \%80).} The percentage of originally dead neurons in the selected paths of different methods reported for sparsity of 80\%. All paths selected by pruning objective contain originally dead (now active) neurons}
\end{figure}
\begin{figure}[!h]
    \centering
    \includegraphics[width=0.99\columnwidth]{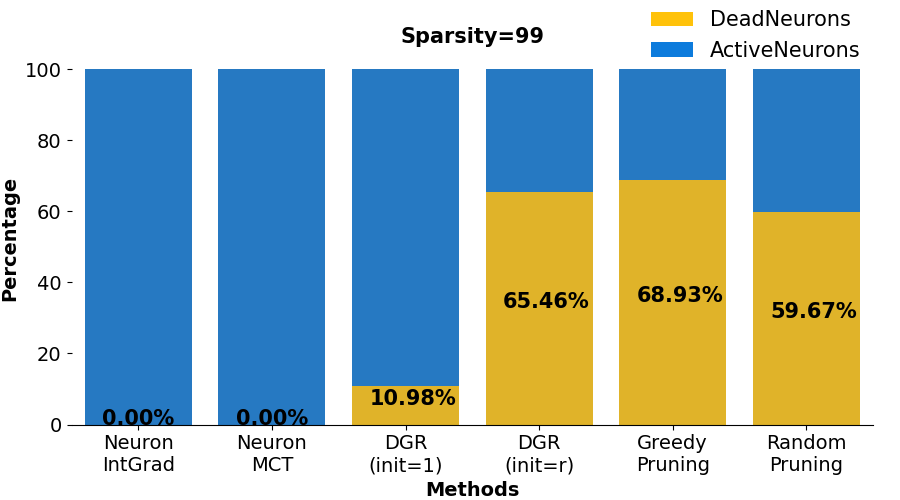}
    \caption{\textbf{Dead Neuron Selection of Pruning Objective (Sparsity \%99).} The percentage of originally dead neurons in the selected paths of different methods reported for sparsity of 99\%. All paths selected by pruning objective contain originally dead (now active) neurons}
\end{figure}
\begin{figure*}[!h]
    \centering
    \includegraphics[width=\linewidth]{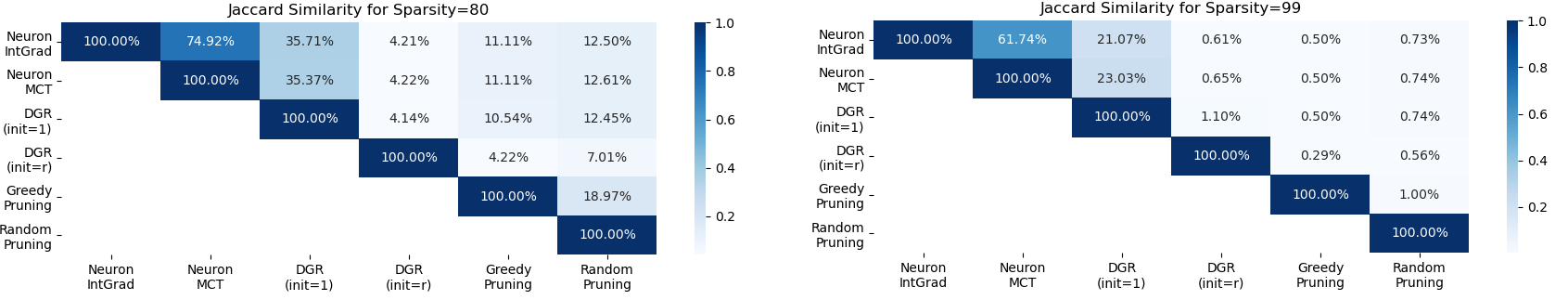}
    \caption{\textbf{Path Analysis - Entire Network (Sparsity 80 \& 99).} Overlap between paths from different methods in entire network. Among the pruning-based methods, only the path selected by DGR(init=1) overlaps with contribution-based methods.}
\end{figure*}
\begin{figure*}[!h]
    \centering
    \includegraphics[width=\linewidth]{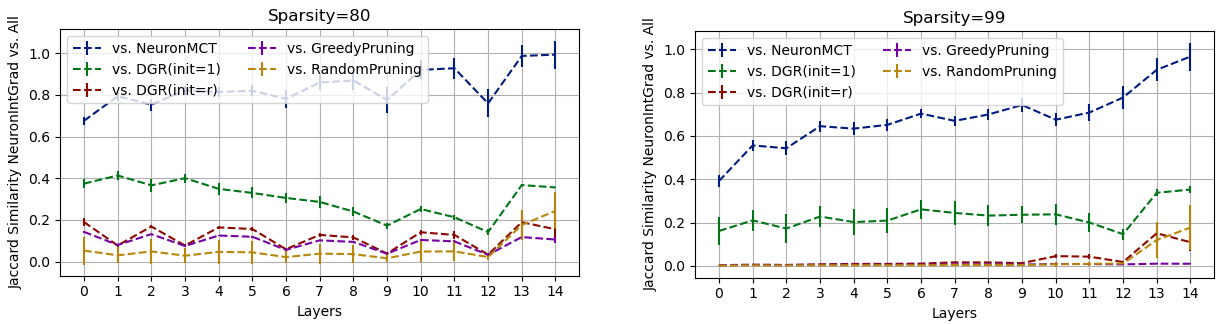}
    \caption{\textbf{Path Analysis - Layerwise (Sparsity 80 \& 99).} Overlap between paths from different methods in different layers of VGG-16. Among the pruning-based methods, only the path selected by DGR(init=1) overlaps with NeuronIntGrad.}
\end{figure*}

\begin{figure*}[!t]
    \centering
    \includegraphics[width=\linewidth]{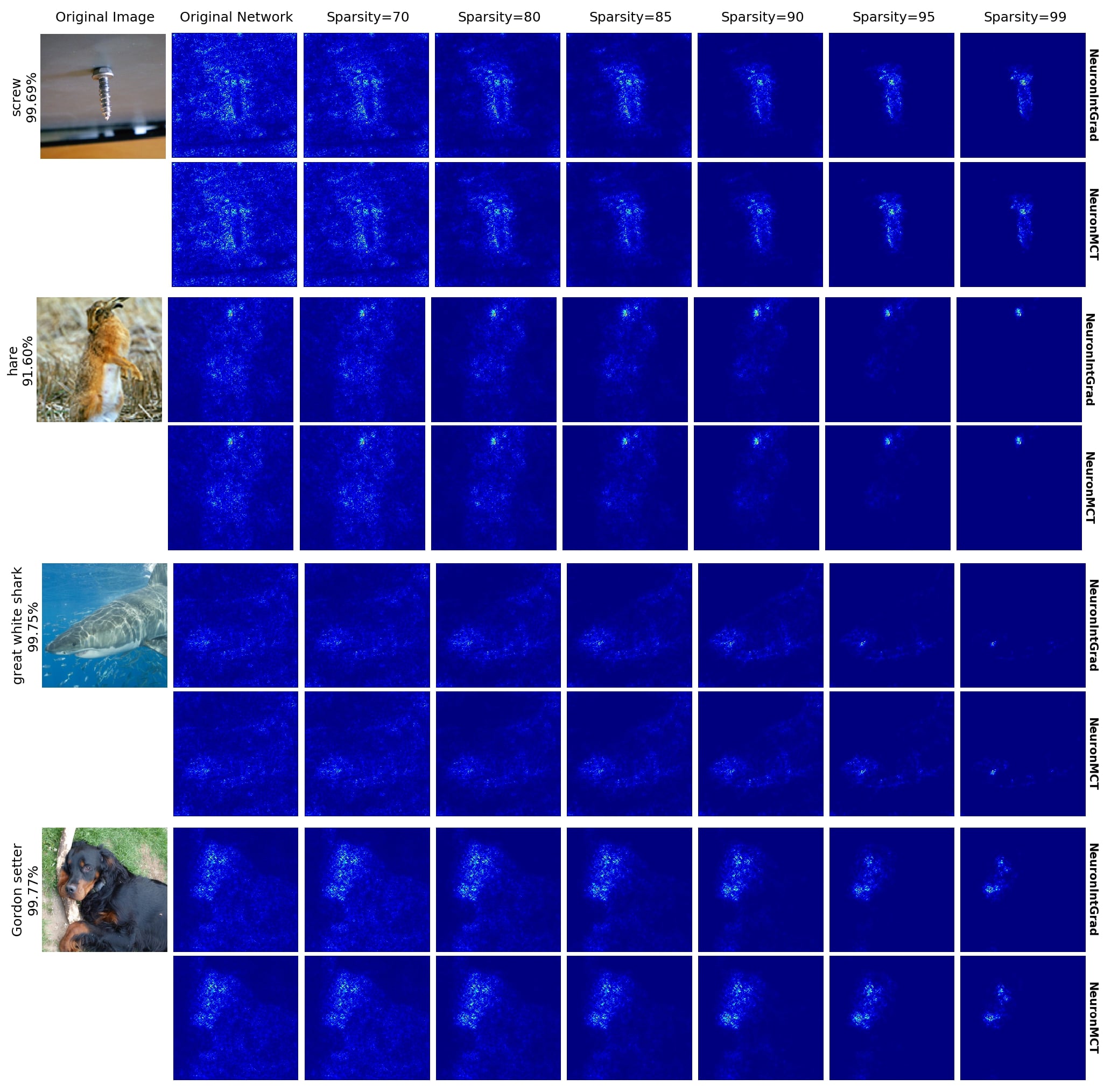}
    \caption{\textbf{Feature Attribution.} The gradients of the locally linear critical paths at different sparsity levels for NeuronIntGrad (top) and NeuronMCT (bottom).}
\end{figure*}
\begin{figure*}[t]
    \centering
    \includegraphics[width=\linewidth]{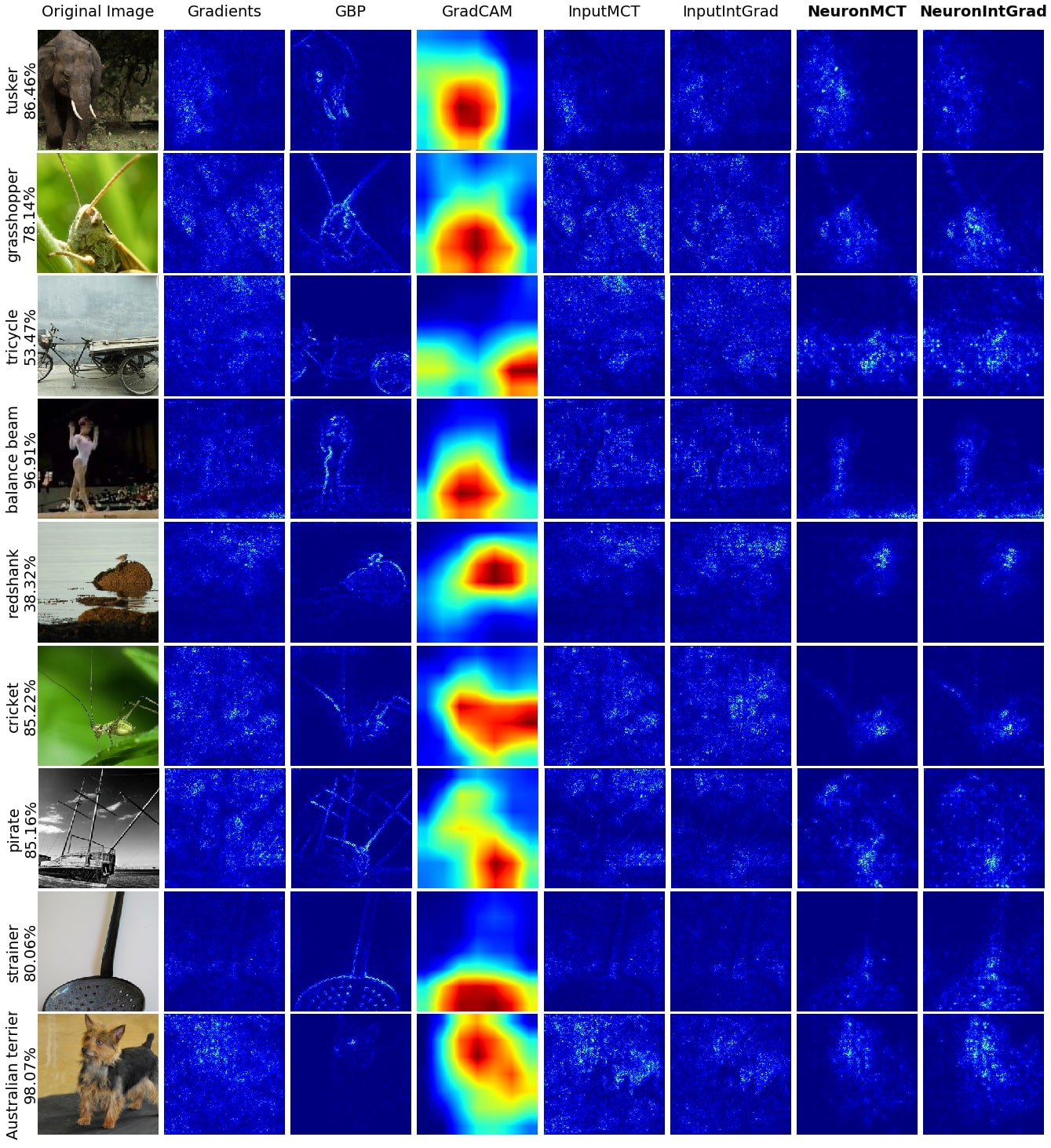}
    \caption{\textbf{Comparison with Feature Attribution Methods.} Comparison between attribution maps dervied our proposed methods (right) vs. gradient-based attribution methods on \textbf{ResNet-50}. Note the improvement of integrated gradients on the neurons (NeuronIntGrad) over integrated gradients on input (InputIntGrad).}
\end{figure*}

\begin{figure*}[t]
    \centering
    \includegraphics[width=\linewidth]{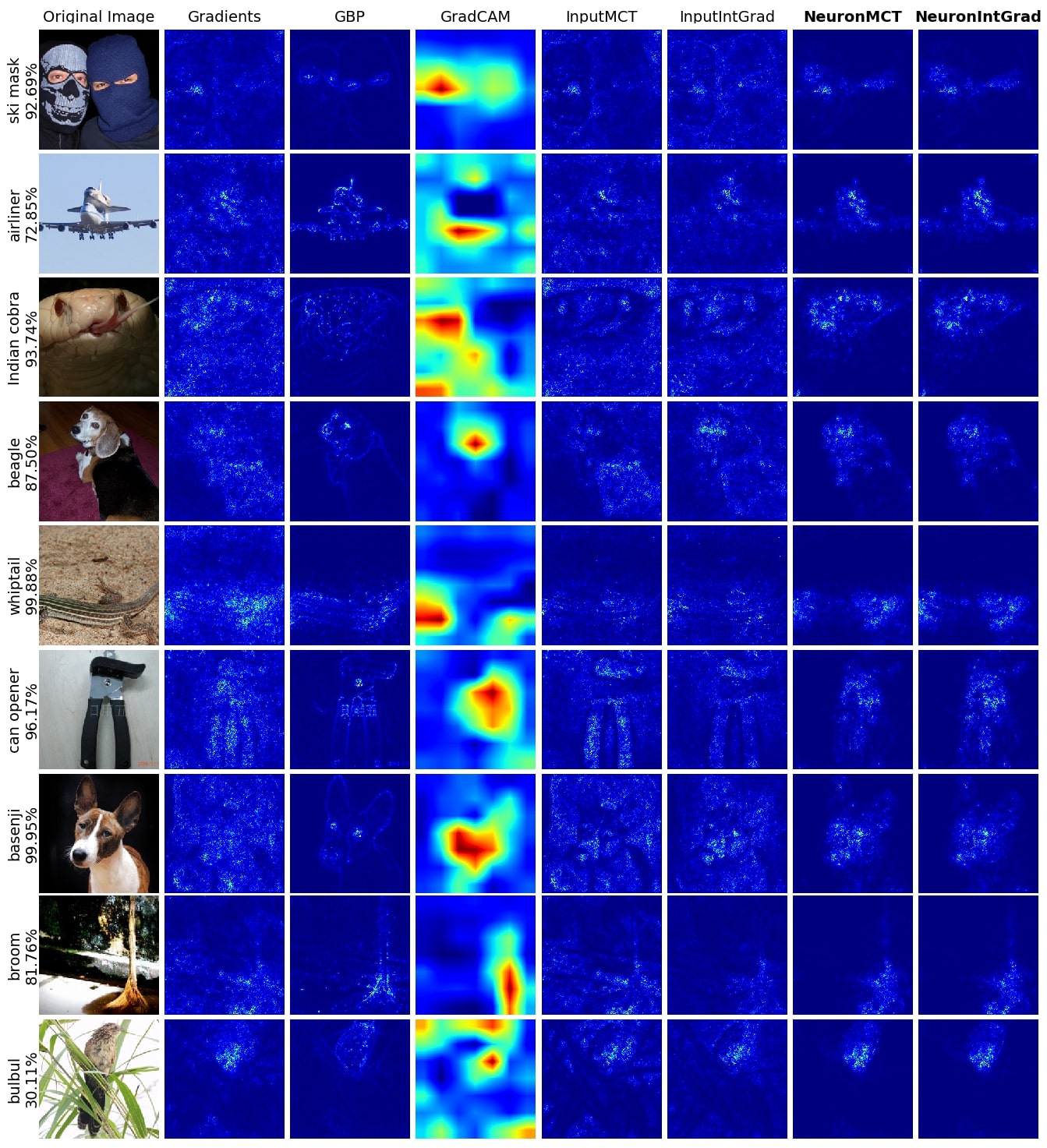}
    \caption{\textbf{Comparison with Feature Attribution Methods.} Comparison between attribution maps dervied our proposed methods (right) vs. gradient-based attribution methods on \textbf{VGG-16}. Note the improvement of integrated gradients on the neurons (NeuronIntGrad) over integrated gradients on input (InputIntGrad).}
\end{figure*}

\end{document}